\documentclass[10pt,letterpaper]{article}
\usepackage[top=0.85in,left=2.75in,footskip=0.75in]{geometry}

\usepackage{amsmath,amssymb}

\usepackage{changepage}

\usepackage[utf8x]{inputenc}

\usepackage{textcomp,marvosym}

\usepackage{cite}

\usepackage{nameref,hyperref}

\usepackage[right]{lineno}

\usepackage{microtype}
\DisableLigatures[f]{encoding = *, family = * }

\usepackage[table]{xcolor}

\usepackage{array}

\newcolumntype{+}{!{\vrule width 2pt}}

\newlength\savedwidth

\raggedright
\usepackage[aboveskip=1pt,labelfont=bf,labelsep=period,justification=raggedright,singlelinecheck=off]{caption}

\makeatletter
\renewcommand{\@biblabel}[1]{\quad#1.}
\makeatother

\usepackage{lastpage,fancyhdr,graphicx}
\usepackage{epstopdf}
\fancyheadoffset[L]{2.25in}
\fancyfootoffset[L]{2.25in}
\begin{document}

\vspace*{0.2in}

\begin{flushleft}
{\Large
\textbf\newline{Adaptive control of underactuated planar pronking hexapod}}

 G\"uner Dil\c{s}ad Er\textsuperscript{1,*},
Mustafa Mert Ankarali\textsuperscript{1,2}

\bigskip
\textbf{1} Department of Electrical and Electronics Engineering, Middle East Technical University (METU), Ankara, Turkey\\
\textbf{2} Robotics and Artificial Intelligence Technologies Application and Research Center (ROMER), Ankara, Turkey\\
\bigskip

* dilsad.er@metu.edu.tr (G. D. E.)

\end{flushleft}

\section*{Abstract}
Underactuated legged robots depict highly nonlinear and complex dynamical behaviors that create significant challenges in accurately modeling system dynamics using both first principles and system identification approaches. Hence, it makes a more substantial challenge to design stabilizing controllers.  If physical parameters on mathematical models have miscalibrations due to uncertainty in identifying and modeling processes, designed controllers could perform poorly or even result in unstable responses. Moreover, these parameters can certainly change-over-time due to operation and environmental conditions. In that respect, analogous to a living organism modifying its behavior in response to novel conditions, adapting/updating system parameters,  such as spring constant,  to compensate for modeling errors could provide the advantage of constructing a stable gait level controller without needing ``exact'' dynamical parameter values. This paper presents an online, model-based adaptive control approach for an underactuated planar hexapod robot's pronking behavior adopted from antelope species. We show through systematic simulation studies that the adaptive control policy is robust to high levels of parameter uncertainties compared to a non-adaptive model-based dead-beat controller.


\section*{Introduction}

Underactuated legged platforms have various motion capabilities than the wheeled robots that preceded them with added mobility such as running, pronking \cite{Ankarali2011pronking}, flipping \cite{SaranliFlip}, and self-righting \cite{SaranliSelfR}. They use their natural dynamics to reach high performance in terms of speed, efficiency, and robustness. In return for this mobility, complex hardware structure and controller design arise as challenges. Marc Raibert decreased this complexity using dynamic modes of locomotion \cite{Rai} in his runners first. Since 1970, Raibert's work has inspired researchers to establish a large new field of study on legged robots \cite{Gregorio1997, Ahmadi2006, MC, Brown1998}. 

If second-order dynamics are appropriately designed and tuned, the model can achieve a wide range of behaviors despite the underactuated nature of many of the legged robotic platforms. However, as the systems become more agile and faster, substantial challenges and problems arise when controlling robot dynamics. Template-based control is an approach to isolate and "independently" control the degrees of freedom relevant to the task  \cite{Peekema2015, Oehlke2016, Saranli2003, Kurtz2021}. In literature, researchers developed template-based control methods for a variety of robots and motions. Saranli and Koditschek applied template-based control for a running hexapod \cite{Saranli2003}. 
Oehlke et al. used template-based hopping control in their research on a bio-inspired segmented robotic leg \cite{Oehlke2016}. Peekema introduced the template-based control of the bipedal robot ATRIAS in his work \cite{Peekema2015}. Furthermore, in recent studies, Kurtz et al. proposed a template-based whole-body controller and simulated it on the 30-DOF Valkyrie humanoid model \cite{Kurtz2021} .

RHex is a structurally simple and yet highly mobile hexapod robot with a single rotary actuator on each hip
\cite{Saranli2001Rhex}. There are contralateral legs that can be used in synchrony for some behaviors. This paper concentrates on the pronking behavior of the RHex platform \cite{Saranli2001Rhex}. 

\begin{figure}[!h]
    \centering
    \vspace{-25mm}
    \includegraphics[width=\textwidth]{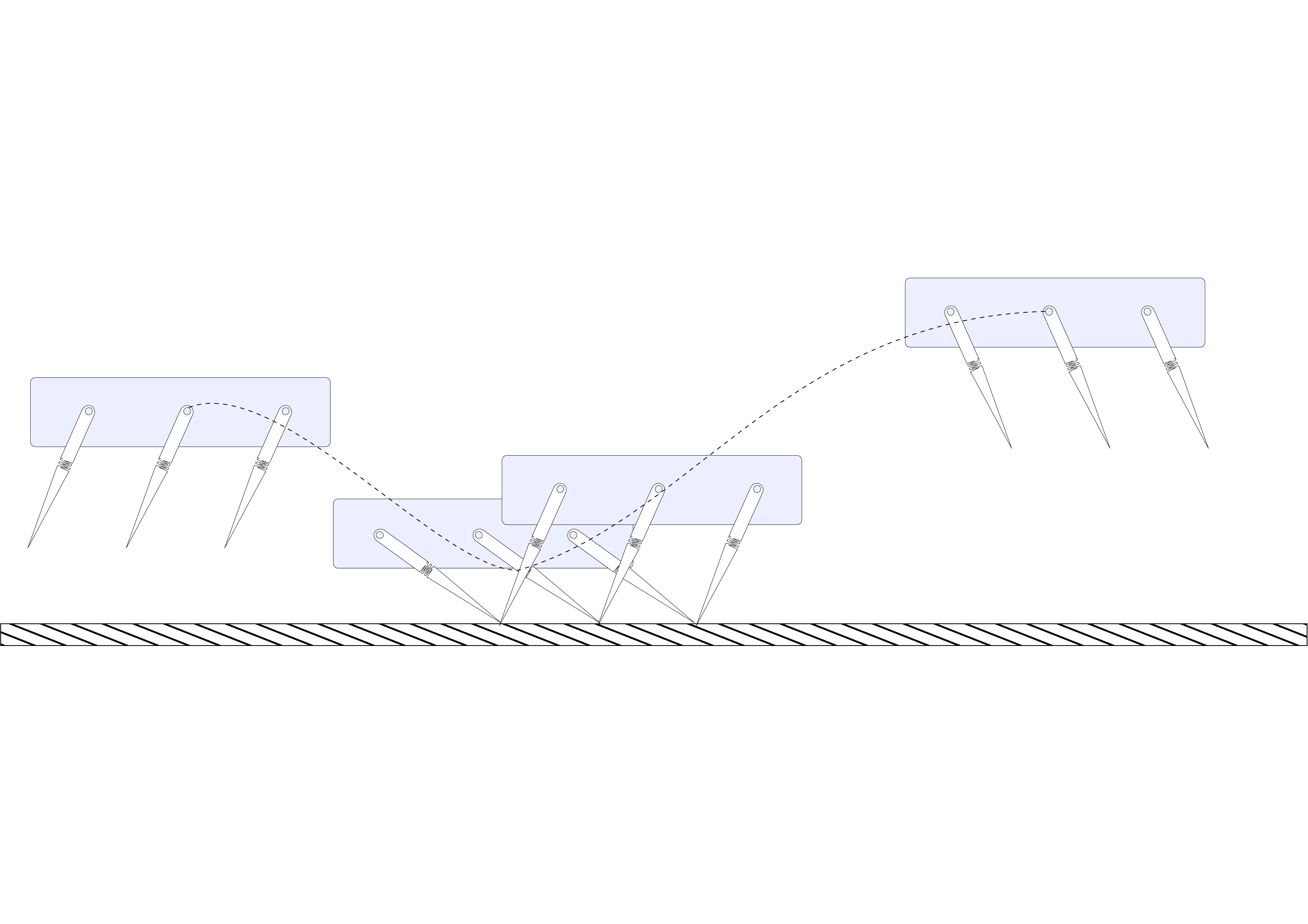}    \vspace{-25mm}

    \caption{{\bf Pronking behavior of planar hexapedal robot}}
    \label{pronking}
\end{figure}

Pronking behavior represents leaping in the air with an arched back and stiff legs, and it is a gait adopted by legged animals such as springbok or other antelope to show their strength to their predators. It is an example of honest signaling in zoology \cite{FitzGibbon1988}. While pronking, animals use their legs in synchrony, and a flight phase follows the stance phase, as depicted in Fig. \ref{pronking}. Pronking gives robots the advantage of considerable jumping heights requiring little ground contact during locomotion. This advantage appealed to scientists to apply controllers to analyze and perform pronking \cite{Mcmordie2002}. Moreover, this is still a living research topic for robots having different types and numbers of legs \cite{Ankarali2011pronking, Johnson2013b, Chou2015, Tseng2019}. In this paper, we propose a new method to improve the pronking motion of the hexapedal RHex platform.

 Although legged robots are advantageous in agility, complex non-linear dynamics of legged locomotion create substantial challenges to the control policy design. Indeed, the complexity of the problem is much deeper than just dealing with complex non-linear computations to generate control actions. The challenge in designing legged locomotion behavior starts with the designing control target itself and developing associated performance metrics, which are relatively trivial in non-legged robotic platforms, such as wheeled systems and aerial platforms. This fact pushed the researchers to inspire from nature and adopt bio-inspired design and control strategies. Arguably, the most famous example of bio-inspiration in legged locomotion is the adoption of the SLIP model in the design of legged platforms and control policies. Numerous studies reported that the SLIP model (and its variants) could accurately capture the center of dynamics and the rhythmic interchange between kinetic and potential energy in running animals. This discovery led the roboticists to develop a number of robotic systems based on the core mechanical principles of the SLIP model. In connection with these developments, several researchers utilized the SLIP model as a high-level control interface for designing control policies for more complex legged platforms \cite{Wensing2013, Grimes2012THEDO, Vanderborght2011}.

Spring-loaded inverted pendulum (SLIP) model \cite{schwindPhd}, which consists of a point mass attached to a free rotating massless leg equipped with a linear spring-damper pair, is a fundamental and straightforward model for understanding fundamental principles in legged locomotion. It is widely used to describe many kinds of robots' leg behavior and brings the advantage of using existing control methods through analytical return maps \cite{geyerMap2005, Ankarali2010AAS}. This paper's proposed scheme is based on analytical approximations to SLIP dynamics \cite{Ankarali2010AAS}, which will be briefly discussed later. Following the introduction of the extended SLIP model with torque actuation at the hip (SLIP-T), the model named Slimpod is developed as a simpler model of hexapedal pronking robot and utilized with an embedded dead-beat controller to realize the pronking motion \cite{Ankarali2011pronking}. 

The aforementioned embedded controller is sensitive to the changes inside the approximate analytical map. Parameters inside the approximate map are fixed but might have deviations from their measured/estimated values on some variables, such as the position of the center of mass, stiffness of the spring, and damping, etc. These discrepancies may result from many reasons, such as measurement errors, corrosion, dirt, and fatigue. In this context, model parameters could adapt to internal or external changes to sustain a stable and robust pronking motion, which is the fundamental starting point of this paper.  

Adaptation to external influence is a broad and essential subject for robotics studies. Especially in physical systems, many sources lead the identification of inner system parameters' be inaccurate. Even if the measurement accuracy is nearly perfect during the initial calibration phase, it can still be necessary to update the parameters automatically as time goes and/or the control task changes.

In literature, there are many examples where the adaptive control framework is utilized as an adaptive locomotion control \cite{Faigl2019, Zhu2016} and for interlimb coordination \cite{Aoi2017}. Besides, some precedents combining with other approaches borrowed from other engineering applications exist, such as adaptive control of a legged robot using artificial neural network  \cite{Helferty1989}, or a combination of evolutionary and adaptive control strategies for a quadruped robot \cite{Massi2019}.  

Adaptive control strategies centers around different approaches on legged robots \cite{Aoi2017,Faigl2019,Helferty1989,Massi2019,Uyanik2011adaptive}. In our system, unlike the previous work in literature \cite{Uyanik2011adaptive}, the intention is not an accurate system identification or an estimation of the environmental effects \cite{Miller2013}. The main goal is to enhance the controller performance for varied circumstances by adding an extra layer to the structure. This strategy evocates the term adaptability for biological organisms \cite{Bateson2017}. We want our robots to inherit the organism's ability to modify their behavior in response to novel conditions. This ability may lead to a possible change in neural control circuits, while meaning a change in inner controller parameters in our robotic systems. Also, prior adaptive controller studies \cite{Miller2013, Uyanik2011adaptive} had focused on simple template models. Different from the previous work in literature, we have a more complex anchor multi-legged model. 

An unknown plant's adaptive control can be carried out by directly adjusting control parameters in a feedback loop based on the error between plan and model outputs, known as direct control. An alternative method is to estimate the plant parameters and to adjust the control parameters based on such estimates described as indirect control \cite{textbook}. Proposed adaptation indirectly affects the dead-beat controller output by amending the chosen parameter in the approximate map used in the embedded dead-beat controller.

Motivated by the previous work in literature \cite{Ankarali2011pronking}, this paper presents an adaptive control method for a pronking hexapedal robot with a spring-loaded inverted pendulum template-based controller. We expect the controller with our proposed adaptation scheme to work better than classical dead-beat controllers thanks to reducing the modeling error. In some sense, adaptive controllers correct approximation errors and, consequentially, provide better tracking on desired height and desired velocity. 

The organization of the paper will be as follows. Model dynamics and control section presents the basis model SLIP and its variations. Adaptive control of Slimpod section introduces the structure of the indirect adaptive control utilized on the objective system. Performance analysis part addresses simulation results, comparisons, and stability analysis. Finally, section 5 closes the work with a conclusion.

\section*{Model Dynamics and Control}

As previously stated in introduction section the proposed method is implemented on a validated planar hexapod model; namely, Slimpod \cite{Ankarali2011pronking}. This section presents an overview of underlying models for the application.

\subsection*{Dynamics of SLIP Template}

\begin{figure}[!htbp]
   \centering
    \includegraphics[width=0.6\textwidth]{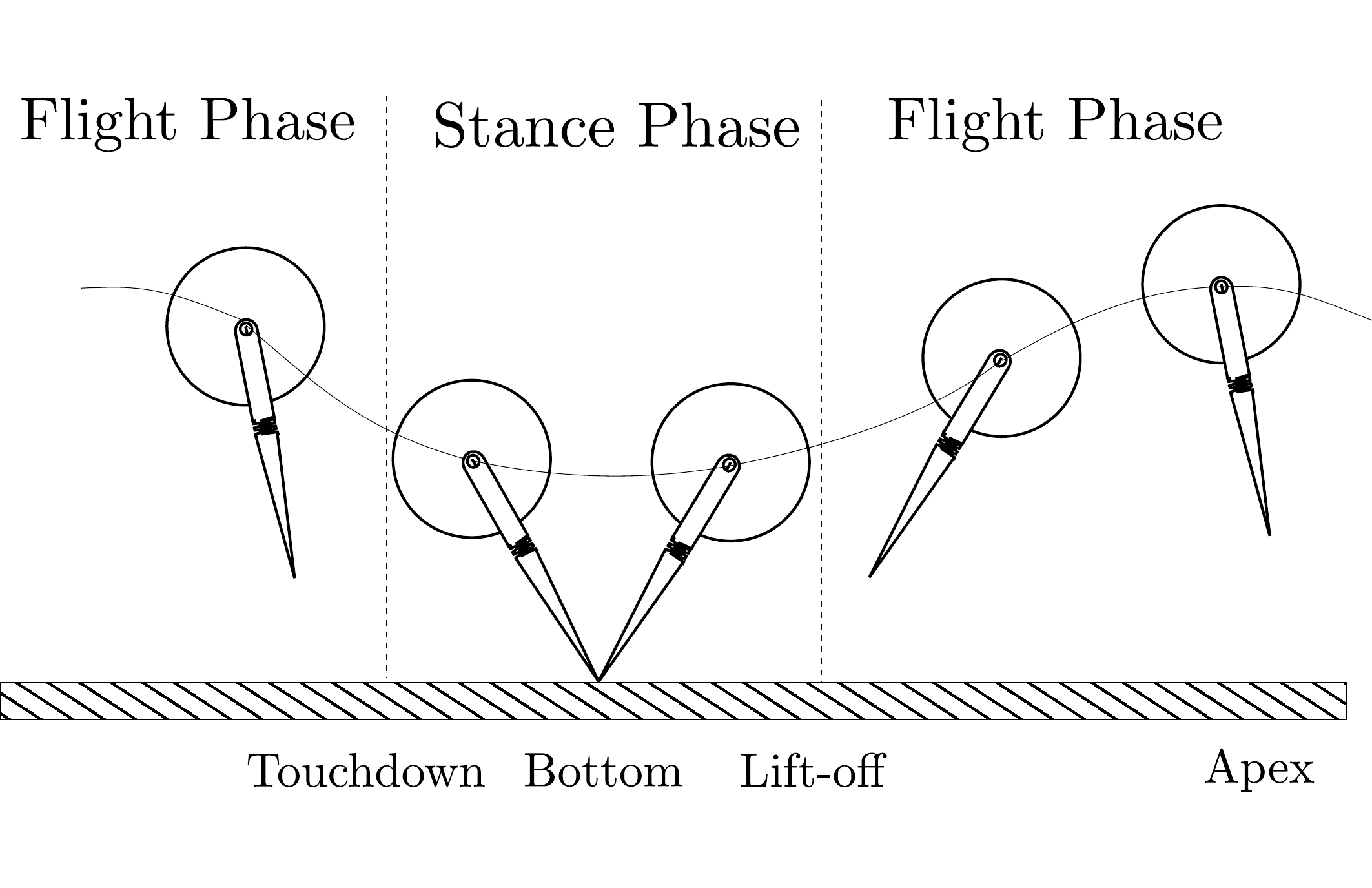}
    \caption{{\bf Locomotion of SLIP template}}
    \label{slip}
\end{figure}

\begin{figure}[!htbp]
    \centering
 \vspace{-10mm}
     \includegraphics[width=0.6\textwidth]{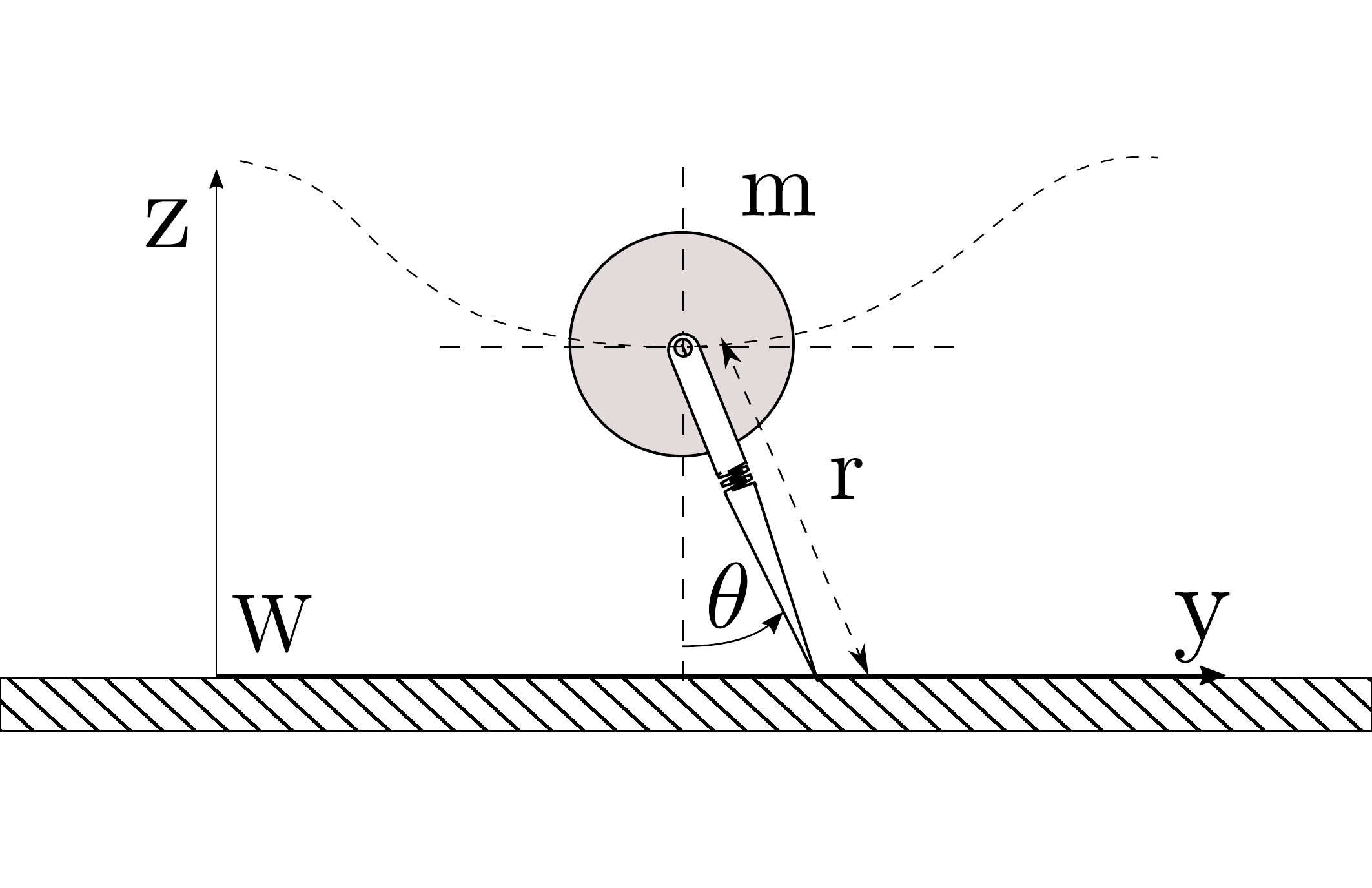}
    \caption{{\bf SLIP template}}
    \label{sliptemp}
\end{figure}

This section refers to the prior SLIP model in order to build a template for the controller. Throughout locomotion, the SLIP model has successive phases named stance and flight. The discrete transitions between those phases are called events, and four important events are touchdown, bottom, lift-off, and apex, as shown in Fig. \ref{slip}. Touchdown and lift-off represent the events on the transition from flight to stance, stance to flight, respectively, and apex represents the event at the point where vertical speed equals zero. Apex state includes the height and horizontal velocity (${}^{a}_{}X^{}_{}=[z,\dot{y}]$). Fig. \ref{sliptemp} depicts the SLIP model with mass m, leg length r, and leg angle $\theta$.

Flight and stance dynamics of SLIP model \cite{Ankarali2011pronking, SaranliHumanoid} obtained from Euler-Lagrange formulation can be written as

Flight:
\begin{align}
\Ddot{y}&=0\\
\Ddot{z}&=-g
\end{align}

Stance:
\begin{align}
    m\Ddot{r}&=mr\dot{\theta}^2+k(l_0-r)-mgcos\theta -b\dot{r}\\
    \Ddot{\theta}&=\frac{d}{dt}(mr^2\dot{\theta})+mgrsin\theta
\end{align}
 
where y and z are the horizontal and vertical positions of the point mass in flight, and $r$ and $\theta$ are the leg length and leg angle in the stance phase.  $k$ and $b$ indicates the leg spring stiffness and leg damping in the formulation, respectively. As can be deduced from the system equations, during flight, the system has a simple projectile trajectory, while in the stance phase, the dynamics become more complex and, more importantly, non-integrable. This non-integrability issue is critical and will appear further in later sections.

The core aim is to design an adaptive controller for the pronking behavior of the RHex platform. So, we need some intermediate models to tie in with the hexapedal RHex model.
The transitional model named SLIP-T \cite{Ankarali2011pronking} builds the necessary connection between SLIP and the Slimpod. SLIP-T differs from SLIP with a single motor at the hip, allowing us to control the input torque instead of radial leg actuation. Additionally, it is a more capable model to project actuator limits. The stance dynamics of SLIP-T have been controlled through the active embedding of ideal SLIP. The embedding controller's primary goal is to find appropriate hip controls to force the dynamics of SLIP-T to the dynamics of simple SLIP. Details of the embedding can be found on Ankarali's paper \cite{Ankarali2011pronking}, which is beyond the scope of the present work.

\subsection*{Dynamics of Slimpod Model}\label{dynSlim}

\begin{figure}[!htbp]
\centering
     \includegraphics[width=0.6\textwidth]{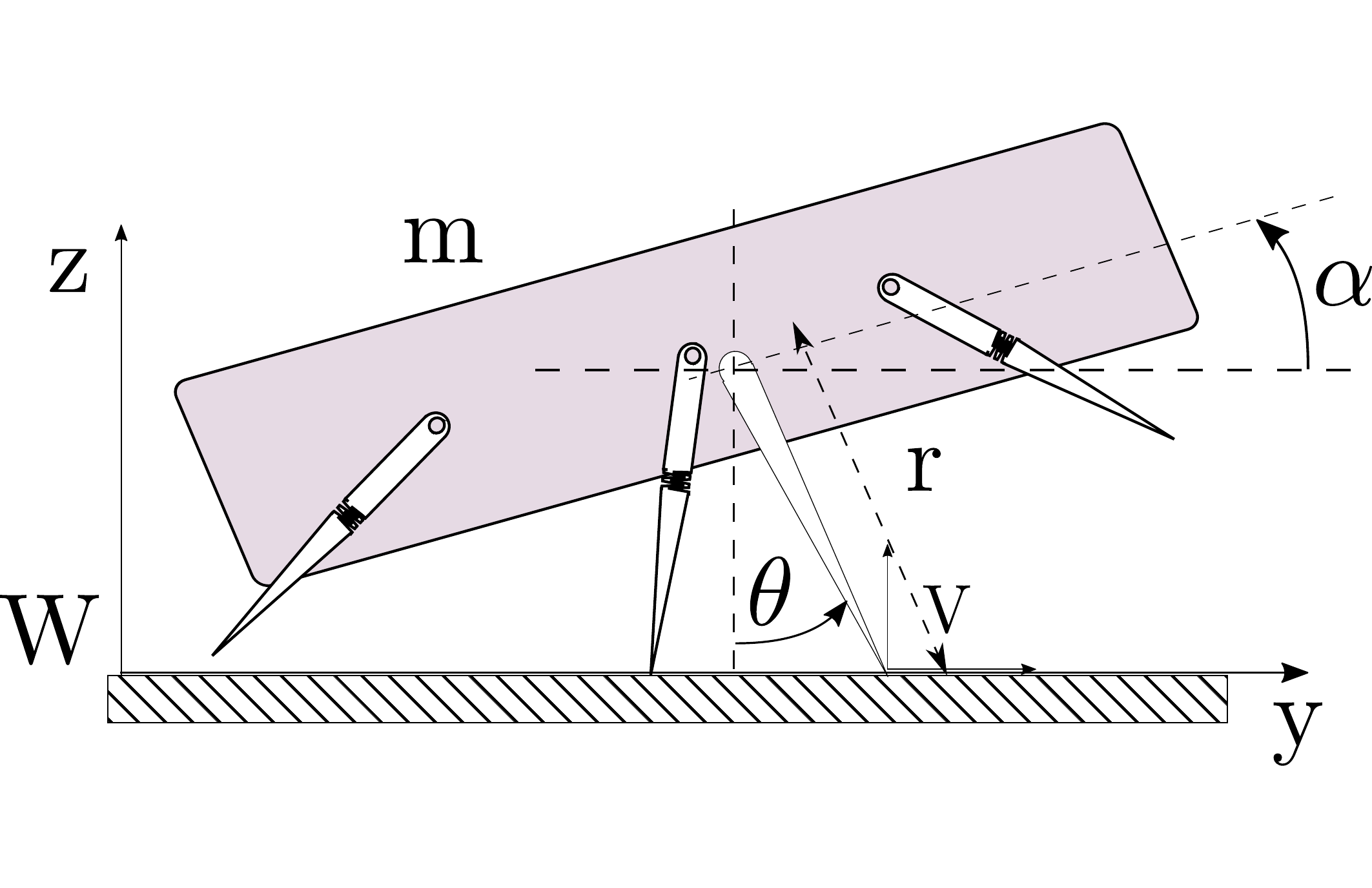}
    \caption{{\bf Slimpod, a planar dynamic model underlying hexapedal RHex robot.} Unlike the generic drawing in this figure, legs are used synchronously during objective pronking motion.}
    \label{slimpod}
\end{figure}

The hexapedal model consists of a body and six legs in contralateral pairs; each has a rotary actuator on its hips. Ankarali utilized a saggital planar model \cite{Ankarali2011pronking} using the leg pairs synchronically for pronking behavior. The planar Slimpod model (\cite{saranliSimsect, saranliPhd}), illustrated in Fig. \ref{slimpod}, allows us to design a feedback controller.

The Slimpod model consists of a rigid body with inertia I and mass m and three legs representing contralateral pairs of RHex, each with a controllable torque. The legs are considered massless during the stance phase, and each leg has a spring with stiffness k\textsubscript{i} and viscous damping coefficient b\textsubscript{i}. Also, a virtual leg is defined (illustrated in Fig.\ref{slimpod}), extending from the body center of mass to a stationary point on the ground.
A flight controller on a lower level drives all legs to their required positions. This drive is based on SLIP control decisions by solving kinematic equations for all legs. This inner controller realizes the desired control inputs by the placement of the virtual leg. After the placement of this virtual toe, the stance controller mimics ideal SLIP dynamics by choosing proper hip torque inputs for each leg of the Slimpod model.

\subsection*{Gait-Level Template-Based Control}

In this paper, we choose the template-based approximate dead-beat control strategy introduced by Ankarali \cite{Ankarali2011pronking} as our baseline (non-adaptive) control strategy. 
Slimpod has a gait level, embedded spring-mass hopper template dead-beat controller to reach the desired apex state. The gait level behavior is summarized through the Poincaré section of its trajectories at each apex point since running is a nonlinear rhythmic motion. The touchdown angle, leg lengths at touchdown, and lift-off are adjusted to achieve the desired apex state. The hyperplane, called the Poincaré section, can be considered passing through apex points where vertical velocity is zero ($\dot{z}=0$).
The time-independent relation between two successive intersections can be defined as a Poincaré map interpreted as
\begin{align}
    f({}^{a}_{}X^{}_{n}) = {}^{a}_{}X^{}_{n+1}
\end{align}

Poincaré map, and analytical approximations proposed by Ankarali et al. \cite{Ankarali2010AAS} allow us to define a discrete return map $f$, and approximate return map indicated as $\hat{f}$, respectively. Note that the discrete return map $f$ has all the information about the physical system parameters accurately while $\hat{f}$ knows these parameters provided in the modeling step, initially. 

\begin{align}
  \hat{f}({}^{a}_{}X^{}_{n},u_n,\hat{p}_n)&={}^{a}_{}\hat{X}^{}_{n+1}
\end{align}

\begin{figure}[!htbp]
    \centering
     \vspace{-10mm} \includegraphics[width=0.9\textwidth]{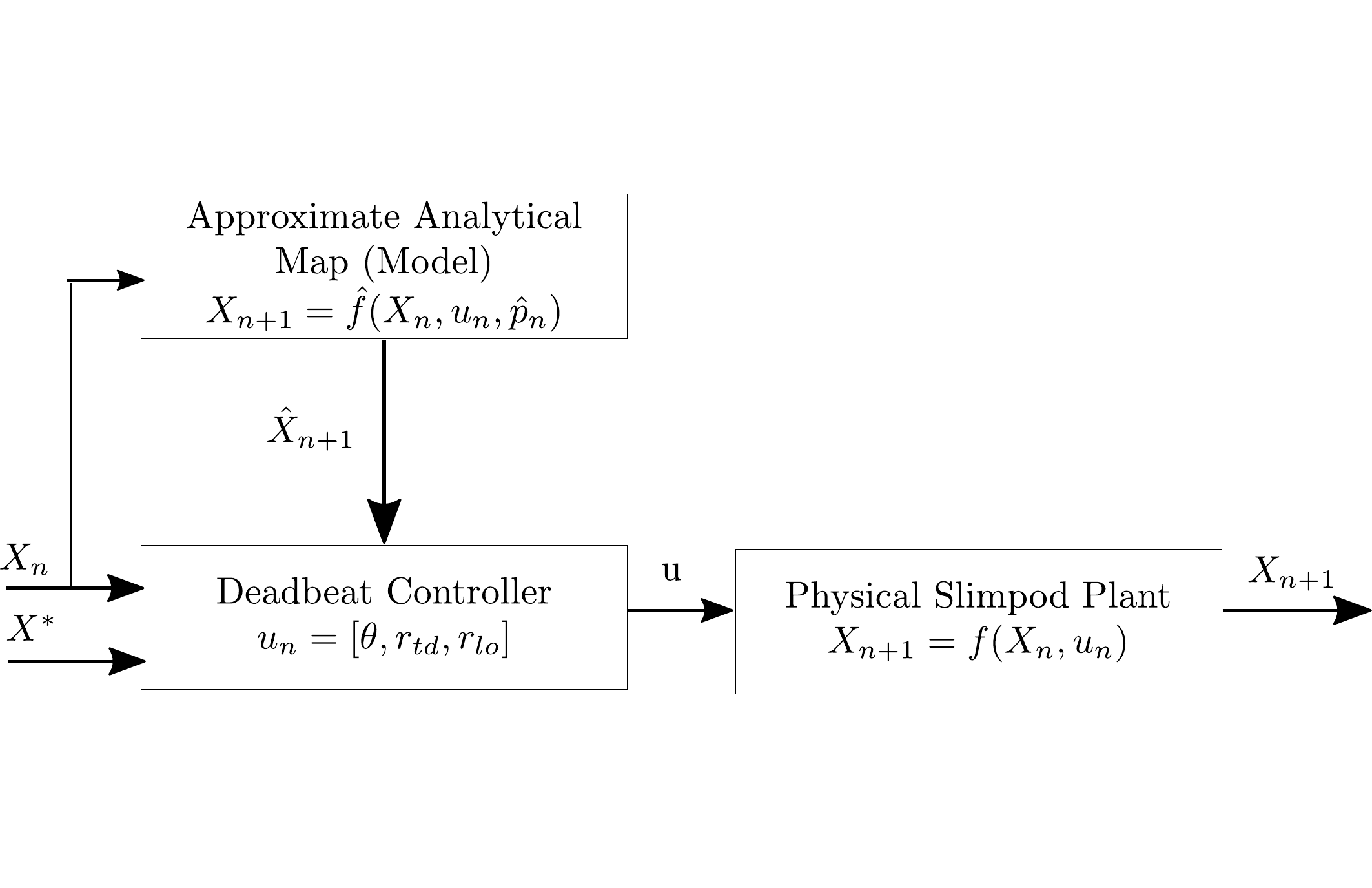}\vspace{-10mm}
    \caption{{\bf Gait-level open-loop dead-beat control scheme.} The baseline controller that is vulnerable to the parameter discrepancies inside Approximate Analytical Map block}
    \label{DBDiagram}
\end{figure}

The dead-beat controller relies on this approximate return map. In simple terms, the map takes the current state and the next touchdown angle and outputs the next state. Dead-beat controller seeks for touchdown angle and leg lengths at touchdown and liftoff ($u=[\theta, r_{td}, r_{lo}]$) with given current apex state (${}^{a}_{}X^{}_{n}=[\dot{y_a}, z_a]$) in order to achieve desired apex height and horizontal velocity (${}^{a}_{}X^{*}_{}=[\dot{y_a}^* z_a^*]$) in one step, through optimization on the map. Indeed, this analytical map includes the system parameters that we should regulate, according to this paper's argument. A simple diagram for dead-beat stride control is given in Fig. \ref{DBDiagram}.

In the literature, previously designed gait-level controllers do not consider the miscalibration or time-varying physical parameter case. On the contrary, they assume a perfect match for the physical quantities in the existing system and the approximate map, making their analytical calculations for control input accordingly. However, the physical parameters of the approximate analytical map inside the baseline controller may be uncertain and time-varying due to various reasons, such as variation in the environment or given task, effects of corrosion, and dust. These distortions and deformations cause extra difficulty in the control of the autonomous system. Also, the measurement error itself arises as an issue, especially for stiffness and damping values. Since the controller performance is closely related to the accuracy of the approximate predictive map, there can be many sources for discrepancies that cause the previous controllers to fail. Therefore, we should adjust the controller according to the error in objective states in order to bring the system to a more realistic side. So, we introduced an adaptive control scheme as an approach to deal with the miscalibration problem.
    
The adopted approximate dead-beat controller utilizes an approximate analytical map based on known system parameters to optimize the control input. If those parameters are miscalibrated, the dead-beat controller eventually will make inaccurate predictions and decisions. Moreover, even if there is a perfect match between all the parameters on the map and the actual system parameters, there will always be some discrepancies between the predictions of the map and actual system outputs due to the approximate nature of the analytical map.  

Dynamics of pronking behavior of planar hexapod robot during phases of toe contact (i.e., stance) \cite{Ankarali2010AAS} are non-integrable under the effect of gravity \cite{poincareChaos}. SLIP dynamics, and also slimpod dynamics, during the stance phase, are related to the restricted three-body problem \cite{SaranliHumanoid}. Restricted three-body problems do not admit to a closed-form solution \cite{poincareChaos}. Therefore, unlike many other adaptive controller applications, \cite{IARC}, constructing a Lyapunov function and deriving adaptive laws for dynamics for pronking motion with a hexapod robot is not reasonably possible. 

\section*{Adaptive Control of Slimpod}

This chapter addresses the proposed indirect adaptive dead-beat controller embedded on the planar hexapedal system model Slimpod. Initially, we proposed the control scheme. The work is followed by the error definition and the parameter update strategy. 

\begin{figure}[!htb]
    \centering
       \vspace{-10mm} \includegraphics[width=0.9\textwidth]{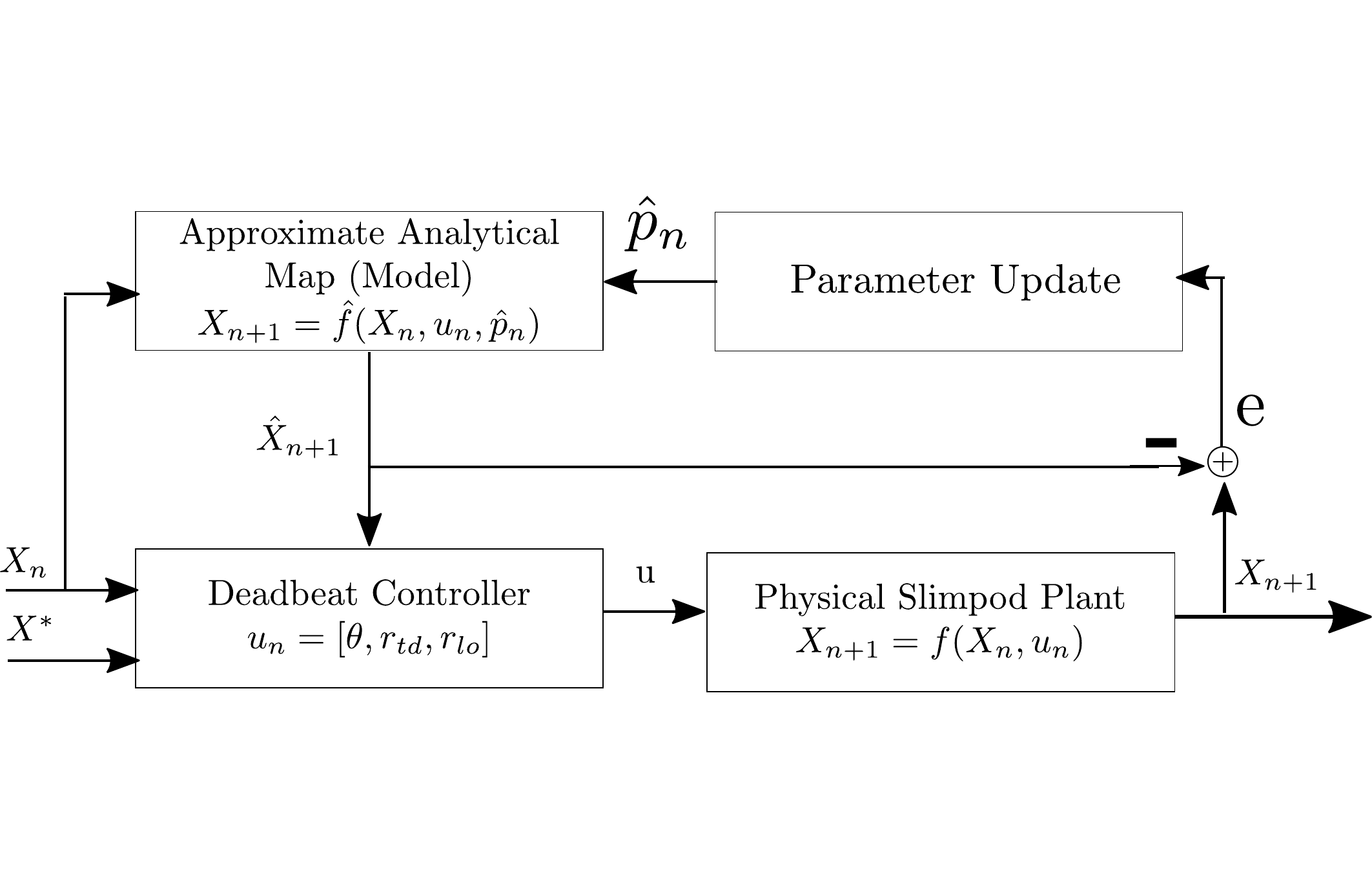}
  \vspace{-10mm}
    \caption{{\bf Adaptive control strategy}}
    \label{AdaptiveDiagram}
\end{figure}

We chose to implement indirect adaptive control because we have a dead-beat controller affected by the objective system parameter and calculates the plant input. Fig. \ref{AdaptiveDiagram} depicts the proposed controller scheme.

\begin{figure*}[h]
 \begin{adjustwidth}{-2.25in}{0in} 
    \centering
    \includegraphics[width=1.5\textwidth]{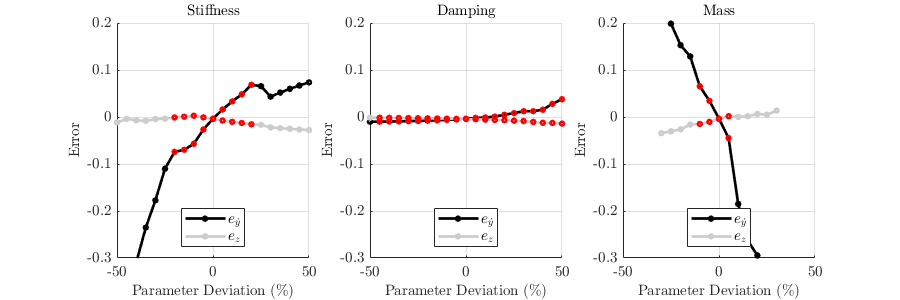}
     \end{adjustwidth}
     \vspace{3mm}
    \caption{{\bf State errors with respect to percentage deformation}}
    \label{dev}
\end{figure*}

As explained under dynamics of slimpod model subsection, Poincaré map, and analytical approximations proposed by Ankarali \cite{Ankarali2010AAS} allow us to define a discrete return map $f$, and approximate return map indicated as $\hat{f}$, respectively. Noting that the approximate return map is constructed based on the inaccurate parameter estimates for the spring's stiffness, the prediction error is also given as
\begin{equation}
\begin{aligned}
e&:= {}^{a}_{}X^{}_{n+1}-{}^{a}_{}\hat{X}^{}_{n+1}\\&=f({}^{a}_{}X^{}_{n},u_n)-\hat{f}({}^{a}_{}X^{}_{n},u_n,\hat{p}_n)\label{error}
\end{aligned}
\end{equation}
The adaptive controller should ensure this error approaches zero. That means the parameter update strategy should satisfy the condition that the model output and the actual output would be equal without disturbing the system's stability.
\begin{align}
lim_{n \rightarrow \infty} ({}^{a}_{}X^{}_{n+1}-{}^{a}_{}\hat{X}^{}_{n+1})=0
\end{align}
To decide how to update our parameters, we need to investigate the impact of miscalibration on our system. In this context, we experimented with three different parameters. Fig. \ref{dev} shows the resulting errors with respect to percentage deviations from ``true'' values. Red marked points represent the experiments when the system can reach a fixed point. The target scenario is to drive both the state error values $e_z$ and their derivatives, e\textsubscript{$\dot{y}$}, to zero. 

As a result of these experiments, chosen parameter and corresponding errors need to have a linear relation to proving the system's stability through linearized system matrices \cite{ModelRed, Hartman1960}. However, deviations in damping (Fig. \ref{dev}) have more of a quadratic behavior, and horizontal velocity state error e\textsubscript{$\dot{y}$} is negative for all deviation amounts regardless of its sign. Therefore, damping is not a proper candidate to adapt in order to reduce absolute error.

Pursuing a consistent notation with the previous study of Ankarali \cite{Ankarali2011pronking}, we adopted a dimensionless formulation for the SLIP model and its variations. These dimensionless expressions eliminate inessential parameters and advance powerful ways of conducting our simulations as well as adapting our controllers to the real physical systems. The SLIP model's dimensionless flight and stance dynamics obtained from the Euler-Lagrange formulation are applied to the dynamics. Representation in dimensionless coordinates converts the physical stiffness k\textsubscript{s} to the dimensionless version r\textsubscript{s}  with $r_s=k_s(l_0/(mg))$ statement \cite{Ankarali2011pronking}, l\textsubscript{0} being the leg length in rest. Therefore, physical individual leg stiffness k\textsubscript{s} and physical body mass $m$ affects the leg stiffness r\textsubscript{s} in dimensionless coordinates. Due to dimensionless coordinates inside the controller, stiffness and mass have similar but symmetric effects on state errors, as observed in Fig. \ref{dev}. Suppose we only study the points regarding the ability to reach a fixed point. In that case, we can infer that deviations in stiffness and mass values inside the approximate analytical map have a linear relation with the state's error value, while deviated damping's effect is nonlinear yet stable.

Taking into consideration that stiffness and mass have both linear effects, we have two candidates. Measurement of the mass value does not require disassembling all the legs and some specific equipment. Mass calibration is more accessible than the stiffness constant calibration, even if the robot has a payload. Hence, we chose stiffness over mass to update. This analysis is our starting point to define our parameter update strategy.

As mentioned before in model dynamics and control section, because the restricted three-body problem does not admit to a closed-form solution, constructing a Lyapunov function and deriving adaptive laws for dynamics of pronking motion with a hexapod robot is not reasonably possible.

MIT rule is the key to remedy this obstacle of nonexistent closed-form solutions \cite{Astrom1995}. The state error is defined as in Eqn. \eqref{error}. The objective parameter $k$ must be updated so that the loss function $V(k)$ in Eqn. \eqref{V} is minimized.
\begin{align}
    V(k)=\frac{1}{2}e^2 \label{V}
\end{align}
The MIT Rule makes $V(k)$ small by changing k in the direction of the negative gradient of $V(k)$.
\begin{equation}
\begin{aligned}
    &\frac{\partial V(k)}{\partial k}= e\frac{\partial e}{\partial k}\\
    \dot{k}=-\gamma &\frac{\partial V(k)}{\partial k}= -\gamma e \frac{\partial e}{\partial k}
\end{aligned}
\end{equation}
$\gamma$ is a positive constant representing the adaptive gain in further steps. Also, we can extract the sensitivity derivative $\frac{\partial e}{\partial k}$ from the Fig. \ref{dev}.

In the literature, the corrective parameter adjustment strategy adopted from the MRAC method \cite{UyanikMS, Uyanik2011adaptive} is very similar to how estimation methods such as Kalman filters use innovation on sensory measurement to perform state updates. Based on all these relations and inspiring from previous work on control related to the spring-mass hopper \cite{Ankarali2011pronking}, which is the basis of our embedded template base controller, we propose a parameter update strategy as
\begin{align}
\hat{p}_{n+1}=\hat{p}_n-K_e*X_n*e
\end{align}
where K\textsubscript{e} is a gain coefficient used to tune convergence of parameter values and regulate the oscillations, note that convergence behavior is strongly related to the adaptive gain K\textsubscript{e}. We will base our implementations on this strategy. Stating the adaptive law leaves us with the question about the system's stability with the adaptive controller. We will discuss the stability issue in stability analysis subsection later.

\section*{Performance Analysis}

This section presents simulation results for reference apex state tracking introduced in adaptive control of slimpod section using different percentage error conditions on two physical parameters, stiffness and damping, and the approximate map itself.  We provided comparisons on adaptive and non-adaptive controller structures to show that the proposed adaptive controller is more capable of producing stable and controllable pronking motion in the case of parametric miscalibration. 

We run all simulations on MATLAB utilizing a hybrid dynamical simulation toolkit based on SimSect \cite{saranliSimsect} previously verified on RHex \cite{saranliPhd}, including necessary additions about slimpod model \cite{Ankarali2011pronking}.
As the solver, we chose a variable-step, variable-order solver based on the numerical differentiation formulas of orders 1 to 5.

Basically, we conducted two types of experiments according to two control schemes in Fig. \ref{DBDiagram} and \ref{AdaptiveDiagram}, and we compare the simulation results in terms of achieving the desired apex state and steady-state error for the given desired state. We ran simulations with various apex goals ${}^{a}_{}X^{*}_{}$ and percentage of miscalibration errors. 

The ranges for the dynamic parameters for the Slimpod model, which apply to a wide range of parameter combinations due to our dimensionless formulation inside the code, detailed in \cite{AnkaraliMS}, were chosen to closely match the physical SensoRHex robot to ensure future applicability of our results to an experimental implementation and given as in Table \ref{table:params}. We plot states $z$ and $\dot{y}$ from the dimensionless group in the following simulation results.


\begin{table}[h]
\caption{\label{table:params}{\bf Parameters of the Slimpod Model}}
\begin{tabular}{||c|c|c|c|c||}
\hline
    Quantity&Symbol & Value & Unit  \\
    \hline
    Body mass & m & 9 & kg \\
    Leg stiffness & k & 2000 & N/m \\ 
    Leg damping & b & 12 & Nm/s\\
    Rest leg length & l & 0.175 & m \\
    Desired height & $z^*$ &0.195 &m \\
    Desired velocity & $y^*$ & 1.6 & m/s\\
   \hline  
\end{tabular}
\end{table}

\begin{figure}[h]
     \vspace{1mm}
 \begin{adjustwidth}{-2.25in}{0in} 
    \centering
      \includegraphics[width=1.5\textwidth]{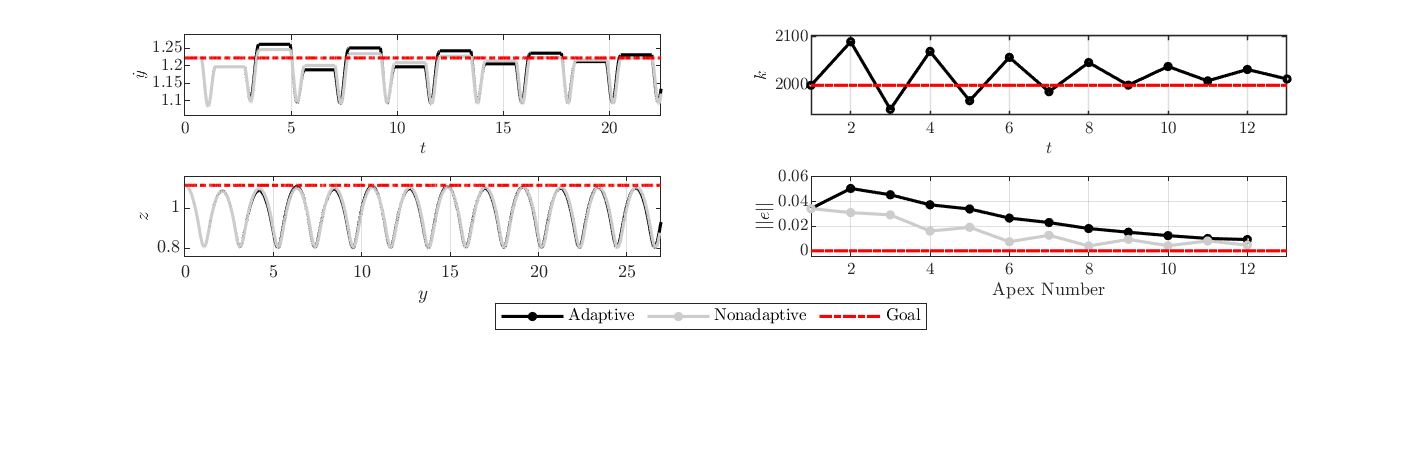}
        \end{adjustwidth}
     \vspace{-8mm}
    \caption{{\bf System response if there is no miscalibration}}
    \label{0}
\end{figure}

Primarily, we performed the simulations, as shown in Fig. \ref{0}, when the system parameters fully match with the controller parameters. From this experiment, we can conclude that even if there is no miscalibration, the adaptive controller reduces the error between model output and actual plant output by updating the $k$ value. This oscillation on the k value leads to fading oscillations on horizontal velocity tracking performance. However, system response seems to become more aggressive, caused by the high parameter update gain.

\begin{figure}[h]
     \vspace{1mm}
 \begin{adjustwidth}{-2.25in}{0in} 
    \centering
  \includegraphics[width=1.5\textwidth]{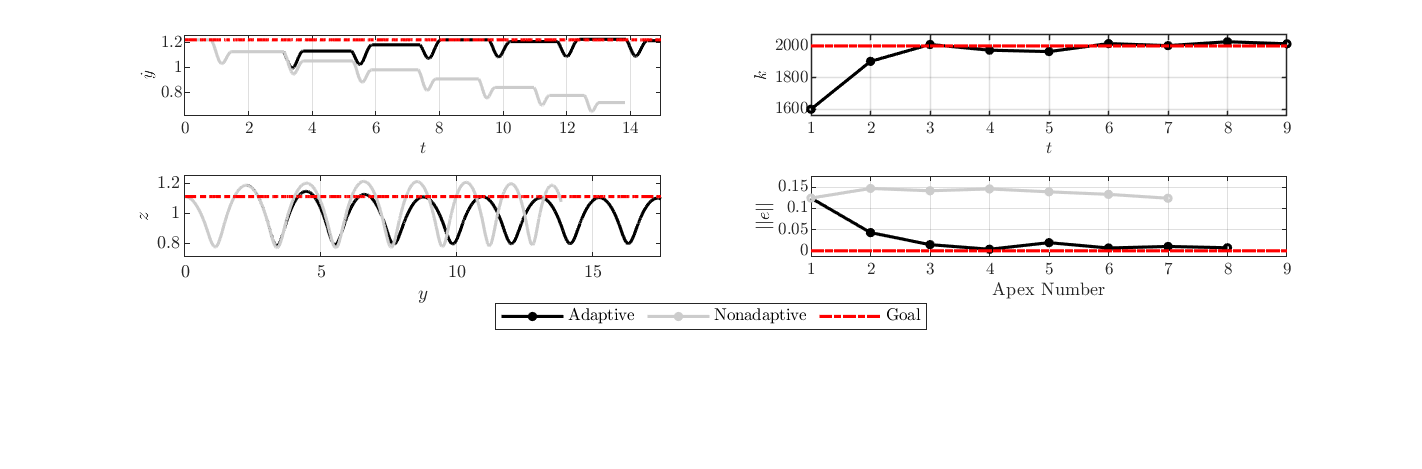}
              \end{adjustwidth}
     \vspace{-8mm}
    \caption{{\bf System response if stiffness on the map is 20\% less than the system value} }
    \label{s-2}
\end{figure}

To signify the effect of the proposed adaptation scheme, we conducted another experiment for the case of a miscalibrated stiffness value and depicted response in Fig. \ref{s-2}. As expected, using the non-adaptive controller with miscalibrated parameters results in large steady-state errors. In addition, after some time, unstable behavior causes the robot to turn upside down. On the other hand, the proposed adaptation scheme clearly reduces the error and promises improved performance. 

\begin{figure}[h]
     \vspace{1mm}
 \begin{adjustwidth}{-2.25in}{0in} 
    \centering
       \includegraphics[width=1.5\textwidth]{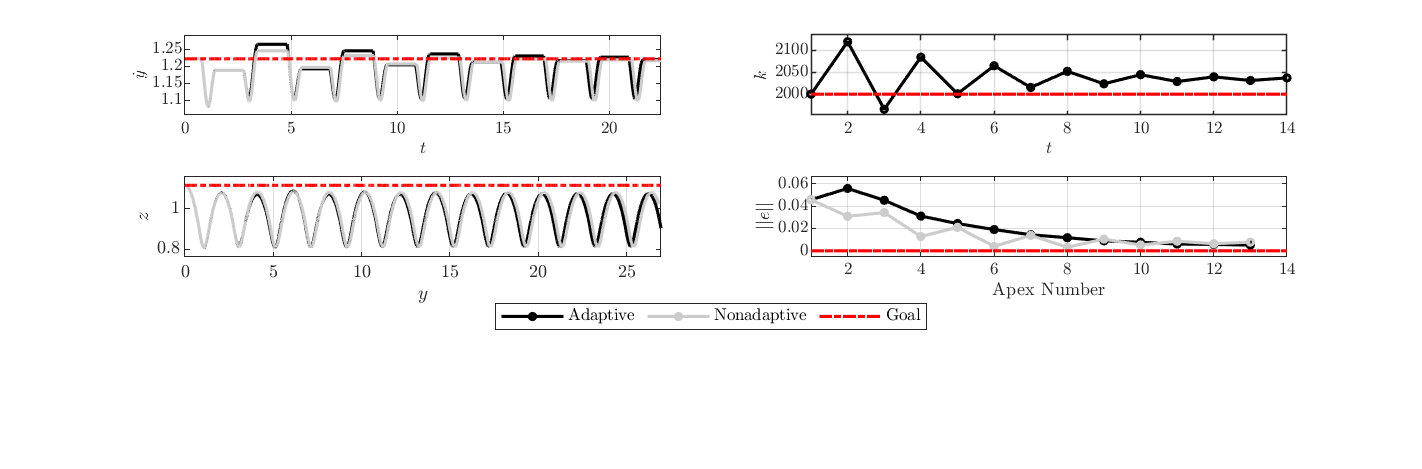}
        \end{adjustwidth}
     \vspace{-8mm}
    \caption{{\bf System response if viscous damping on the map is 10\% less than the system value}}
    \label{d-1}
\end{figure}

We repeated the simulation in Fig. \ref{d-1} for the miscalibrated damping value and noted that stiffness adaptation leads the robot to recover the tracking properly. Just updating stiffness helps to compensate for the error caused by the miscalibrated damping.

\begin{figure}[h]
     \vspace{1mm}
 \begin{adjustwidth}{-2.25in}{0in} 
    \centering
\includegraphics[width=1.5\textwidth]{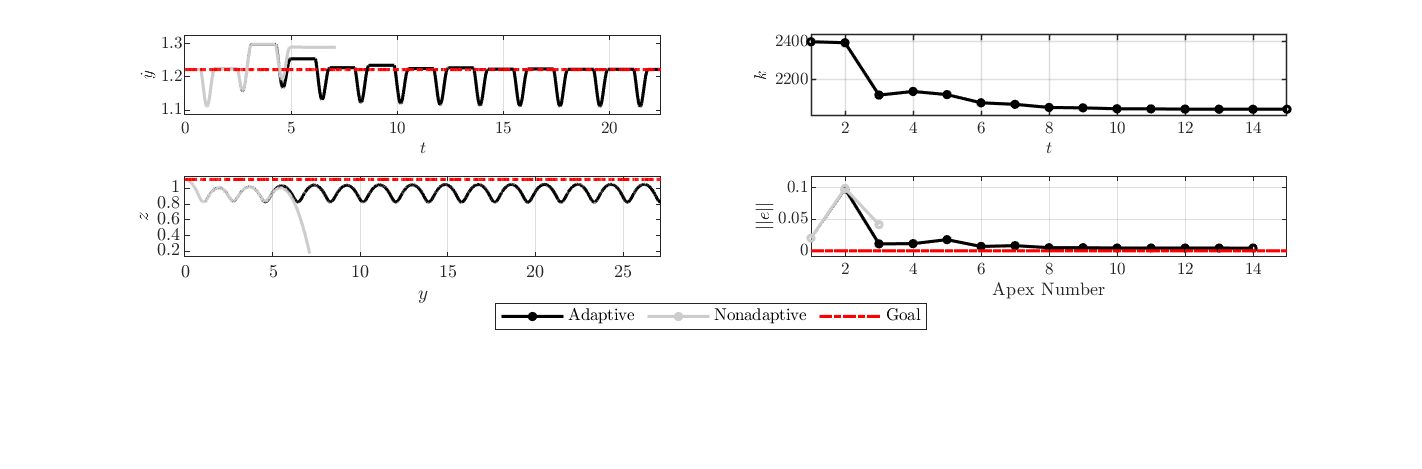} \end{adjustwidth}
     \vspace{-8mm}
    \caption{{\bf System response if stiffness on the map is 20\% and viscous damping is -20\% deviate from the system value}}
    \label{s2d-2}
\end{figure}

In the following simulation, we show the performance for a multiple miscalibration problem in Fig. \ref{s2d-2}. According to these simulation results, when both stiffness and damping parameters are miscalibrated, updating stiffness helps decrease estimation error, and consequentially tracking performance is improved.

\begin{figure}[!h]
     \vspace{1mm}
    \centering
     \begin{adjustwidth}{-2.25in}{0in}
   \includegraphics[width=1.5\textwidth]{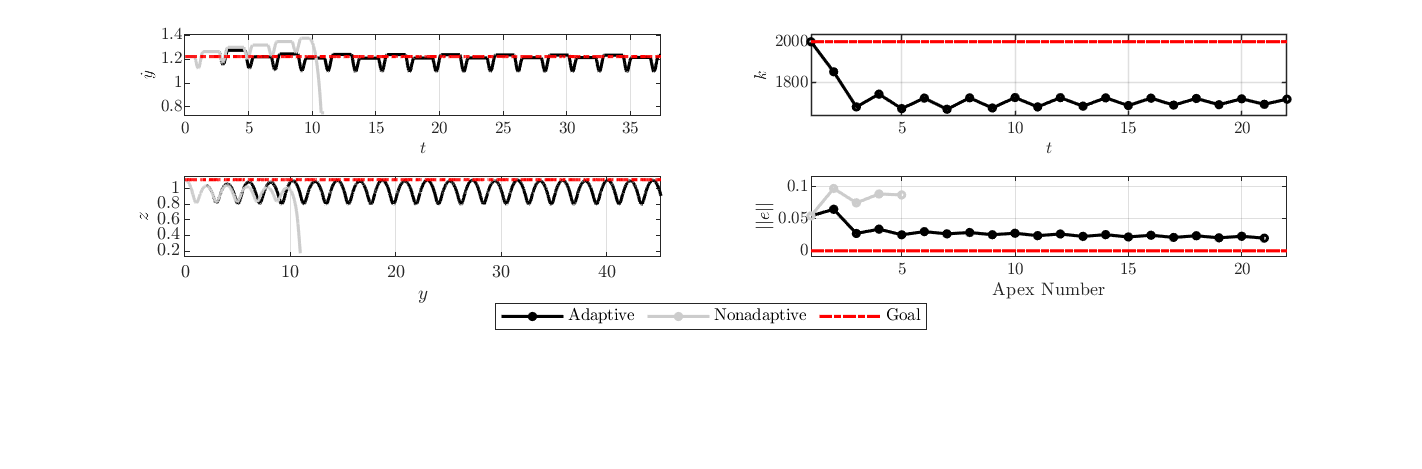}\end{adjustwidth}
     \vspace{-8mm}
    \caption{{\bf System response if the result of approximate map deviates -5\%}} 
    \label{bc}
\end{figure}

\begin{figure}[!h]
      \vspace{1mm}
    \centering
    \begin{adjustwidth}{-2.25in}{0in}
    \includegraphics[width=1.5\textwidth]{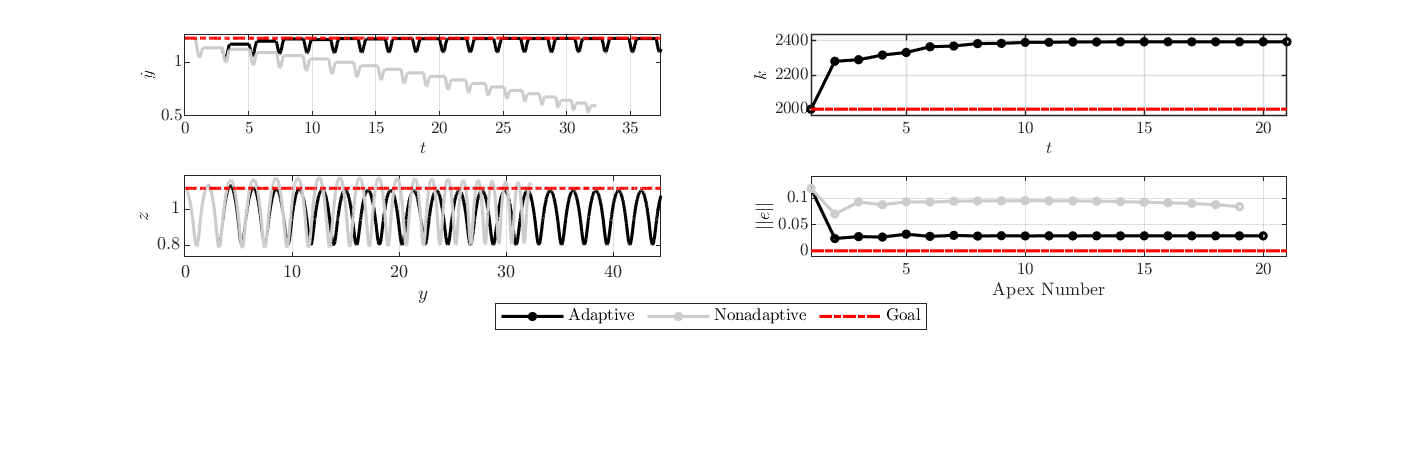}
    \end{adjustwidth}
     \vspace{-8mm}
    \caption{{\bf System response if the result of approximate map deviates +5\% }}
    \label{bc+}
\end{figure}

The final two simulations are devoted to assessing the controller's performance in the case that we directly disturb the output of the approximate map with a constant amount corresponding to 5\% of the desired state values. Resulting responses will be as in Fig. \ref{bc} and \ref{bc+}. According to these figures, if the approximate map's output deviates from its value by 5\% higher or lower, states are drastically affected, making the system unstable. The addition of the adaptation compensates for the error and ensures the system's stability to some extend.

In each case, the adaptive controller updates the stiffness value utilized on the approximate analytical map inside the dead-beat controller and improves tracking performance. Briefly, we may conclude that the proposed adaptive control scheme to adjust stiffness value could confront the miscalibration problem in the planar hexapedal system.

\begin{figure*}[!h]
    \centering
       \begin{adjustwidth}{-2.25in}{0in}
    \includegraphics[width=1.5\textwidth]{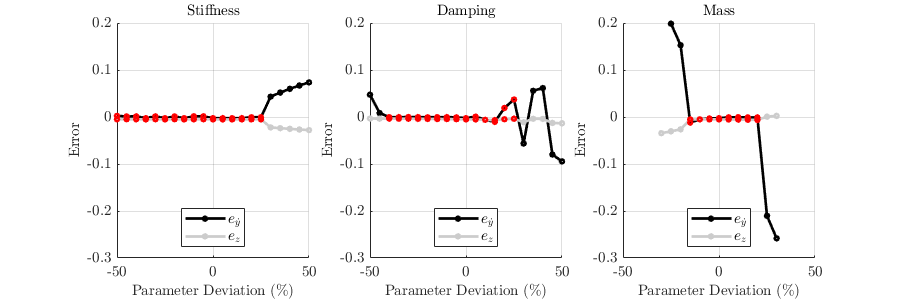}
     \end{adjustwidth}
     \vspace{3mm}
    \caption{{\bf State errors with respect to percentage deformation after the involvement of parameter adaptation scheme for stiffness}}
    \label{Addev}
\end{figure*}

In Fig. \ref{Addev}, we run the parameter adaptation for stiffness together with the dead-beat controller when stiffness, mass, and damping are miscalibrated. In other words, we repeated the experiments in Fig. \ref{dev} with adaptive stiffness. Comparing Fig. \ref{dev} and \ref{Addev}, the systems with parameter adaptation are able to regulate in a wider range of percentage error, driving state errors closer to zero levels.

\newpage
\subsection*{Stability Analysis}\label{stability}

A need arises for developing a way to show the system's stability with an adaptive controller since we cannot develop an adaptive law using standard design procedures in adaptive control \cite{textbook}. Because running is nonlinear, complex periodic motion, the Poincar\'e Map method is used in the analysis. This method intersects a hyperplane with the periodic trajectory of a system with n-dimensional state space. This hyper-plane is called the Poincar\'e section. If the intersection is called X\textsubscript{k}, Poincar\'e Map is defined as $f(X_k) = X_{k + 1}$. Therefore, the relationship between two consecutive intersections can be defined as independent of time. Suppose the map is $f(X^{*})=X^{∗}$, i. e., trajectories intersect with Poincar\'e section on the same point, that point is called a fixed point. It is possible to comment on the system's stability by looking at the local stability at its fixed points \cite{Stride, ModelRed}. 

\begin{figure}[!h]
    \centering
\includegraphics[width=0.6\textwidth]{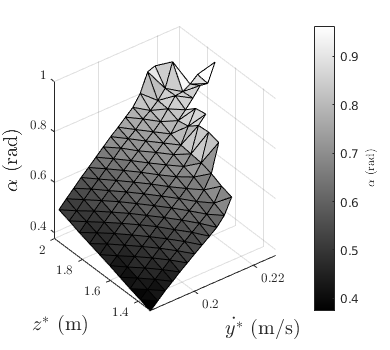}
    \caption{{\bf Variation of the fixed point subspace} by height $z$, horizontal velocity $\dot{y}$ and the body angle $\alpha$}
    \label{fixed_pts}
\end{figure}

Embedded dead-beat controller assures the fixed points in a wide subspace. Fixed point subspace given in Table \ref{table:region} is constructed by considering the gait level controllable region \cite{Ankarali2011pronking} in order to investigate stability. Fixed points are depicted in Fig. \ref{fixed_pts}.


\begin{table}[!h]
\caption{\label{table:region}{\bf Chosen regions for fixed points}}
\begin{tabular}{||c|c||}
\hline
    State &  Physical Values \\
    \hline
    $z^*$ & [0.1850,0.2750] m \\
    $y^*$ & [1.3096,1.9644] m/s\\
   \hline  
\end{tabular}
\end{table}

We analyzed the system's stability resulting from the proposed parameter adjustment strategy based on its linearized system matrix. Study in \cite{ModelRed} supports that hybrid systems' behavior can be reduced to lower-dimensional subsystems near periodic orbits. Therefore, we decide the stability of fixed points by investigating eigenvalues of the numerically calculated linearized system matrix (on Eqn. \eqref{jacob}) \cite{NumMATLAB}. As known from all discrete systems, if all eigenvalues' magnitudes (i.e., $||\lambda_{max}||$) are smaller than 1, the system is stable. 

\begin{align}
\label{jacob}
\hat{J}=\frac{\partial f}{\partial X_k}(X^*)
\end{align}

\begin{figure}[!h]
    \centering
\includegraphics[width=0.6\textwidth]{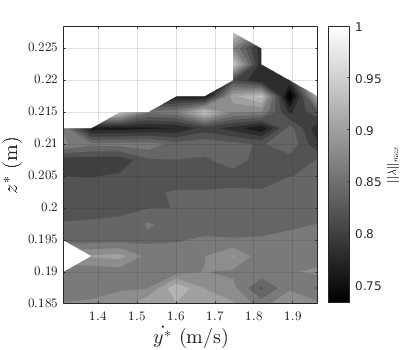}
\caption{{\bf $||\lambda_{max}||$ of linearized system matrices calculated by using the fixed points given in Fig.\ref{fixed_pts}}}
    \label{eigval}
\end{figure}

This linearized system matrix $\hat{J}$ relates the infinitesimal changes in apex state predictions $\partial \hat{X}$ to infinitesimal changes in states. We calculated the linearized system matrices for tracking different height and horizontal velocity pairs. Eigenvalues at those points can be observed in Fig. \ref{eigval}. As depicted, all linearized system matrices have their $||\lambda_{max}||$ inside the unit circle, i.e., their magnitudes are smaller than one, which is the stability criterion for a discrete-time system.

\begin{figure}[!h]
    \centering
 \includegraphics[width=0.6\textwidth]{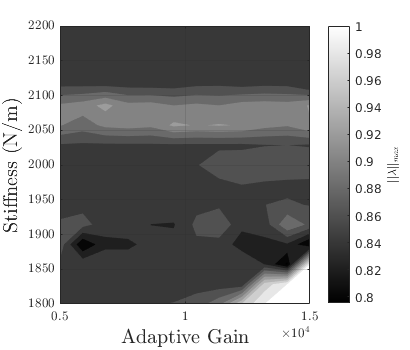}
\caption{{\bf $||\lambda_{max}||$ of linearized system matrices calculated with different stiffness and gain pairs}}
    \label{eigval_stiffness}
\end{figure}

We also investigated linearized numerical stability for different leg stiffness and adaptive gain pairs. Fig \ref{eigval_stiffness} exhibits that our system with adaptive controller remains stable in a wide range of stiffness and gain values.

\newpage
\section*{Conclusion}

In this paper, we proposed an online parameter adaptation scheme for an underactuated legged robot to improve the previously proposed control algorithms' tracking performance and robustness to parametric uncertainties. First, we analyzed the apex state error and parameter relation and defined the parameter update rule accordingly. We then demonstrated the contribution of our additional level to the existing approximate dead-beat controller scheme. In succession, these steps admit to deploying a high-level layer to the existing controller for the Slimpod model.

The formerly developed dead-beat controller's goal is to bring the apex states to desired values. Resultant control input entirely relies on the actual values of the parameters. Inevitably, miscalibrations in the measurements of physical instruments or parameter changes due to environmental effects will create a significant estimation error, so the dead-beat controller's decisions become inaccurate. We added a layer for adaptation to improve the tracking performance as much as possible. Simulation results for constructing a parameter adaptation structure show that adding the parameter update realizes this objective successfully. It reduces the error between model and actual plant outputs and substantially improves control performance relative to existing non-adaptive controllers.

This study extends the relationships between the apex state errors and the legs' physical parameters. The complexity of these relations mainly arises from the coupled effects of the parameters on gait behavior. We inferred how parameter calibration affects the apex states with an experimental investigation. Using the presented approach in this paper, we can now generalize this experimental design procedure to different systems which are not suitable for analytical calculations of indirect adaptive control. In addition, we proved local exponential stability of the closed-loop system under the adaptive rule via computing linearized (numerical) Poincar\'e return map around the emergent fixed-points and checking the associated eigenvalues  \cite{ModelRed,ankarali2014system}. 

Inspiration from the different animals' natural ability for varying gaits draws attention to legged systems. Employing controllers that mimic various motions allows robots to exploit different advantages, such as energy efficiency, large jumping heights, or speeds. Through adaptation, robots can now take a further step to embark on an important biotic characteristic to increase the feasibility of the existing theoretical controllers and practice in more complex places. In the future, we intend to scale this process to different robotic platforms with unmeasurable (directly) inner or environmental parameters. Online adaptive parameter update rule can create a relatively simple yet powerful and practical approach for legged robotic platforms. As a result, the applicability of the theoretical controllers to the actual physical systems will be foreseen to rise significantly.



\begin{thebibliography}{10}

\bibitem{Ankarali2011pronking}
Ankarali MM, Saranli U.
\newblock {Control of underactuated planar pronking through an embedded
  spring-mass Hopper template}.
\newblock Autonomous Robots. 2011;30:217--231.
\newblock doi:{10.1007/s10514-010-9216-x}.

\bibitem{SaranliFlip}
Saranli U, Koditschek D.
\newblock {Design and Analysis of a Flipping Controller for RHex}.
\newblock UM, Ann Arbor, MI; 2010.

\bibitem{SaranliSelfR}
Saranli U, Rizzi A, Koditschek D.
\newblock In: {Multi-Point Contact Models for Dynamic Self-Righting of a
  Hexapod}. vol.~17. Springer; 2005. p. 409--424.

\bibitem{Rai}
Raibert MH.
\newblock {Legged Robots that Balance}.
\newblock MIT Press; 1986.

\bibitem{Gregorio1997}
Gregorio P, Ahmadi M, Buehler M.
\newblock {Design, control, and energetics of an electrically actuated legged
  robot}.
\newblock IEEE Transactions on Systems, Man, and Cybernetics, Part B
  (Cybernetics). 1997;27(4):626--634.
\newblock doi:{10.1109/3477.604106}.

\bibitem{Ahmadi2006}
Ahmadi M, Buehler M.
\newblock {Controlled passive dynamic running experiments with the ARL-monopod
  II}.
\newblock IEEE Transactions on Robotics. 2006;22(5):974--986.
\newblock doi:{10.1109/TRO.2006.878935}.

\bibitem{MC}
McGeer T.
\newblock {Passive bipedal running}.
\newblock Royal Soc B. 1990;213(240):107--134.

\bibitem{Brown1998}
Brown B, Zeglin G.
\newblock {The bow leg hopping robot}.
\newblock In: Proceedings. 1998 IEEE International Conference on Robotics and
  Automation (Cat. No.98CH36146). vol.~1. IEEE; 1998. p. 781--786.

\bibitem{Peekema2015}
Peekema AT.
\newblock {Template-Based Control of the Bipedal Robot ATRIAS} [MastersThesis].
\newblock Oregon State University; 2015.

\bibitem{Oehlke2016}
Oehlke J, Sharbafi MA, Beckerle P, Seyfarth A.
\newblock {Template-based hopping control of a bio-inspired segmented robotic
  leg}.
\newblock Proceedings of the IEEE RAS and EMBS International Conference on
  Biomedical Robotics and Biomechatronics. 2016;2016-July:35--40.
\newblock doi:{10.1109/BIOROB.2016.7523595}.

\bibitem{Saranli2003}
Saranli U, Koditschek DE.
\newblock {Template based control of hexapedal running}.
\newblock In: 2003 IEEE International Conference on Robotics and Automation
  (Cat. No.03CH37422). vol.~1. IEEE; 2003. p. 1374--1379.

\bibitem{Kurtz2021}
Kurtz V, Wensing PM, Lin H.
\newblock {Approximate Simulation for Template-Based Whole-Body Control}.
\newblock IEEE Robotics and Automation Letters. 2021;6(2):558--565.
\newblock doi:{10.1109/LRA.2020.3047794}.

\bibitem{Saranli2001Rhex}
Saranli U, Buehler M, Koditschek D.
\newblock {RHex: A Simple and Highly Mobile Hexapod Robot}.
\newblock The International Journal of Robotics Research. 2001;20(7):616--631.
\newblock doi:{10.1177/02783640122067570}.

\bibitem{FitzGibbon1988}
Fitzgibbon CD, Fanshawe JH.
\newblock {Stotting in Thomson ' s gazelles : an honest signal of condition}.
\newblock Behavioral Ecology and Sociobiology. 1988; p. 69--74.

\bibitem{Mcmordie2002}
Mcmordie D.
\newblock {Towards pronking with a hexapod robot}.
\newblock McGill University, Montreal, Canada; 2002.

\bibitem{Johnson2013b}
Johnson AM, Koditschek DE.
\newblock {Legged self-manipulation}.
\newblock IEEE Access. 2013;1(33):310--334.
\newblock doi:{10.1109/ACCESS.2013.2263192}.

\bibitem{Chou2015}
Chou YC, Huang KJ, Yu WS, Lin PC.
\newblock {Model-based development of leaping in a hexapod robot}.
\newblock IEEE Transactions on Robotics. 2015;31(1):40--54.
\newblock doi:{10.1109/TRO.2014.2376141}.

\bibitem{Tseng2019}
Tseng KY, Lin PC.
\newblock {Development of Leaping/Flipping Behaviors in a Quadruped Robot with
  Passive Compliant Legs}.
\newblock In: 2019 IEEE/ASME International Conference on Advanced Intelligent
  Mechatronics (AIM). vol. 2019-July. IEEE; 2019. p. 364--369.

\bibitem{Wensing2013}
Wensing PM, Orin DE.
\newblock In: Kumar V, Schmiedeler J, Sreenivasan SV, Su HJ, editors. {Control
  of Humanoid Hopping Based on a SLIP Model}. Heidelberg: Springer
  International Publishing; 2013. p. 265--274.

\bibitem{Grimes2012THEDO}
Grimes JA, Hurst JW.
\newblock {The Design of ATRIAS 1.0 a Unique Monopod, Hopping Robot}.
\newblock In: Adaptive Mobile Robotics. WORLD SCIENTIFIC; 2012. p. 548--554.

\bibitem{Vanderborght2011}
Vanderborght B, Tsagarakis NG, {Van Ham} R, Thorson I, Caldwell DG.
\newblock {MACCEPA 2.0: compliant actuator used for energy efficient hopping
  robot Chobino1D}.
\newblock Autonomous Robots. 2011;31(1):55--65.
\newblock doi:{10.1007/s10514-011-9230-7}.

\bibitem{schwindPhd}
Schwind WJ.
\newblock {Spring loaded inverted pendulum running: a plant model}.
\newblock UM, Ann Arbor, MI; 1998.

\bibitem{geyerMap2005}
Geyer H, Seyfarth A, Blickhan R.
\newblock {Spring-mass running: simple approximate solution and application to
  gait stability}.
\newblock Journal of Theoretical Biology. 2005;232(3):315--328.
\newblock doi:{10.1016/j.jtbi.2004.08.015}.

\bibitem{Ankarali2010AAS}
Saranli U, Arslan {\"{O}}, Ankarali MM, Morgul O.
\newblock {Approximate analytic solutions to non-symmetric stance trajectories
  of the passive Spring-Loaded Inverted Pendulum with damping}.
\newblock Nonlinear Dynamics. 2010;62(4):729--742.
\newblock doi:{10.1007/s11071-010-9757-8}.

\bibitem{Faigl2019}
Faigl J, {\v{C}}{\'{i}}{\v{z}}ek P.
\newblock {Adaptive locomotion control of hexapod walking robot for traversing
  rough terrains with position feedback only}.
\newblock Robotics and Autonomous Systems. 2019;116:136--147.
\newblock doi:{10.1016/j.robot.2019.03.008}.

\bibitem{Zhu2016}
Zhu Q, Mao Y, Xiong R, Wu J.
\newblock {Adaptive Torque and Position Control for a Legged Robot Based on a
  Series Elastic Actuator}.
\newblock International Journal of Advanced Robotic Systems. 2016;13(1):26.
\newblock doi:{10.5772/62204}.

\bibitem{Aoi2017}
Aoi S, Manoonpong P, Ambe Y, Matsuno F, W{\"{o}}rg{\"{o}}tter F.
\newblock {Adaptive control strategies for interlimb coordination in legged
  robots: A review}.
\newblock Frontiers in Neurorobotics. 2017;11(AUG):1--21.
\newblock doi:{10.3389/fnbot.2017.00039}.

\bibitem{Helferty1989}
Helferty, Kam.
\newblock {Adaptive control of a legged robot using an artificial neural
  network}.
\newblock In: IEEE International Conference on Systems Engineering. IEEE; 1989.
  p. 165--168.

\bibitem{Massi2019}
Massi E, Vannucci L, Albanese U, Capolei MC, Vandesompele A, Urbain G, et~al.
\newblock {Combining evolutionary and adaptive control strategies for quadruped
  robotic locomotion}.
\newblock Frontiers in Neurorobotics. 2019;13(August):1--19.
\newblock doi:{10.3389/fnbot.2019.00071}.

\bibitem{Uyanik2011adaptive}
Uyanik I, Saranli U, Morgul O.
\newblock {Adaptive control of a spring-mass hopper}.
\newblock In: 2011 IEEE International Conference on Robotics and Automation.
  IEEE; 2011. p. 2138--2143.

\bibitem{Miller2013}
Miller BD, Cartes D, Clark JE.
\newblock {Leg stiffness adaptation for running on unknown terrains}.
\newblock In: IEEE International Conference on Intelligent Robots and Systems;
  2013. p. 5108--5113.

\bibitem{Bateson2017}
Bateson P.
\newblock Adaptability and evolution.
\newblock Interface Focus. 2017;7(5):20160126.
\newblock doi:{10.1098/rsfs.2016.0126}.

\bibitem{textbook}
Narendra KS, Annaswamy AM.
\newblock {Stable Adaptive Systems}.
\newblock USA: Prentice-Hall, Inc.; 1989.

\bibitem{SaranliHumanoid}
Geyer H, Saranli U.
\newblock {Gait Based on the Spring-Loaded Inverted Pendulum}.
\newblock In: Humanoid Robotics: A Reference. Springer; 2018. p. 923--947.

\bibitem{saranliSimsect}
Saranli U.
\newblock {SimSect Hybrid DynamicalSimulation Environment}.
\newblock UM, Ann Arbor, MI: Technical Report CSE-TR-436-00; 2000.

\bibitem{saranliPhd}
Saranli U.
\newblock {Dynamic locomotion with a hexapod robot}.
\newblock UM, Ann Arbor, MI; 2002.

\bibitem{poincareChaos}
Holmes P.
\newblock {Poincar{\'{e}}, celestial mechanics, dynamical-systems theory and
  “chaos”}.
\newblock Physics Reports. 1990;193(3):137--163.
\newblock doi:{10.1016/0370-1573(90)90012-Q}.

\bibitem{IARC}
Mohanty A, Yao B.
\newblock {Indirect Adaptive Robust Control of Hydraulic Manipulators With
  Accurate Parameter Estimates}.
\newblock IEEE Transactions on Control Systems Technology. 2011;19(3):567--575.

\bibitem{ModelRed}
Burden SA, Revzen S, Sastry SS.
\newblock {Model Reduction Near Periodic Orbits of Hybrid Dynamical Systems}.
\newblock IEEE Transactions on Automatic Control. 2015;60(10):2626--2639.
\newblock doi:{10.1109/TAC.2015.2411971}.

\bibitem{Hartman1960}
Hartman P.
\newblock {A lemma in the theory of structural stability of differential
  equations}.
\newblock Proceedings of the American Mathematical Society.
  1960;11(4):610--610.
\newblock doi:{10.1090/S0002-9939-1960-0121542-7}.

\bibitem{Astrom1995}
Astrom KJ, Wittenmark B.
\newblock {Adaptive Control}.
\newblock 2nd ed. Mineola, New York: Dover Publications; 1995.

\bibitem{UyanikMS}
Uyanik I.
\newblock {Adaptive control of a one-legged hopping robot through dynamically
  embedded spring-loaded inverted pendulum template}.
\newblock Bilkent University, TR. Bilkent University, TR; 2011.

\bibitem{AnkaraliMS}
Ankarali MM.
\newblock {Control of hexapedal pronking through a dynamically embedded spring
  loaded inverted pedulum template}.
\newblock Middle East Technical University, TR. Middle East Technical
  University, TR; 2010.

\bibitem{Stride}
Ankarali MM, Saranli U.
\newblock {Stride-to-stride energy regulation for robust self-stability of a
  torque-actuated dissipative spring-mass hopper}.
\newblock Chaos: An Interdisciplinary Journal of Nonlinear Science.
  2010;20(3):033121.
\newblock doi:{10.1063/1.3486803}.

\bibitem{NumMATLAB}
Mathews JH, Fink KD.
\newblock {Numerical methods using MATLAB}.
\newblock USA: Prentice-Hall, Inc.; 1999.

\bibitem{ankarali2014system}
Ankarali MM, Cowan NJ.
\newblock System identification of rhythmic hybrid dynamical systems via
  discrete time harmonic transfer functions.
\newblock In: 53rd IEEE Conference on Decision and Control. IEEE; 2014. p.
  1017--1022.

\end{thebibliography}
\end{document}